\documentclass[letterpaper, 10 pt, conference]{ieeeconf}
\pdfoutput=1

\IEEEoverridecommandlockouts
\usepackage{cite}
\usepackage{amsmath,amssymb,amsfonts}
\usepackage{algorithmic}
\usepackage{graphicx}
\usepackage{textcomp}
\usepackage{xcolor}

\usepackage{dsfont}
\usepackage{algorithm}
\usepackage{subfigure}
\usepackage{booktabs}
\usepackage{balance}
\usepackage{comment}
\usepackage{url}
\usepackage{color}
\usepackage{array}
\usepackage{multirow}

\def\BibTeX{{\rm B\kern-.05em{\sc i\kern-.025em b}\kern-.08em
    T\kern-.1667em\lower.7ex\hbox{E}\kern-.125emX}}
\begin{document}

\title{\LARGE \bf KDD-LOAM: Jointly Learned Keypoint Detector and Descriptors Assisted LiDAR Odometry and Mapping
}

\author{Renlang Huang, Minglei Zhao, Jiming Chen, and Liang Li
\thanks{The authors are with the College of Control Science and Engineering,
        Zhejiang University, Hangzhou, 310027, P. R. China.}
}

\maketitle

\begin{abstract}
Sparse keypoint matching based on distinct 3D feature representations can improve the efficiency and robustness of point cloud registration.
Existing learning-based 3D descriptors and keypoint detectors are either independent or loosely coupled, so they cannot fully adapt to each other.
In this work, we propose a tightly coupled keypoint detector and descriptor (TCKDD) based on a multi-task fully convolutional network with a probabilistic detection loss.
In particular, this self-supervised detection loss fully adapts the keypoint detector to any jointly learned descriptors and benefits the self-supervised learning of descriptors.
Extensive experiments on both indoor and outdoor datasets show that our TCKDD achieves \textit{state-of-the-art} performance in point cloud registration.
Furthermore, we design a keypoint detector and descriptors-assisted LiDAR odometry and mapping framework (KDD-LOAM), whose real-time odometry relies on keypoint descriptor matching-based RANSAC.
The sparse keypoints are further used for efficient scan-to-map registration and mapping.
Experiments on KITTI dataset demonstrate that KDD-LOAM significantly surpasses LOAM and shows competitive performance in odometry.
\end{abstract}

\section{Introduction}
\label{sec:intro}

Point cloud registration is crucial for many robotic and 3D vision applications, such as simultaneous localization and mapping (SLAM)~\cite{behley2018suma} and 3D reconstruction~\cite{zeng20173dmatch}.
Although the classic iterative closest point (ICP) algorithm~\cite{icp1992tpami} can precisely estimate the transformation, it requires an initial guess close to the ground truth and inefficient iterations to establish correct correspondences.
In contrast, sparse feature matching directly establishes reliable correspondences between keypoints with similar descriptors, achieving efficient and robust point cloud registration.

Even though a few hand-crafted 3D keypoint detectors~\cite{chen2007det,zhong2009iss} and descriptors~\cite{rusu2008pfh,rusu2009fpfh,salti2014shot} have been proposed over the years, the performance of 3D-3D point association remains unsatisfactory.
In contrast, the learning-based descriptor is regarded as a promising approach \cite{bai2020d3feat}, which maps the low-level geometric representations to a discriminative feature space.
These descriptors are always learned in a self-supervised manner by maximizing the similarity between corresponding point features and minimizing the similarity between other point pairs.
However, as it is difficult to define and label keypoints, these descriptors usually overlook keypoint detection and randomly sample points for description and matching, thus suffering from several drawbacks. 
First, inefficient oversampling is required to ensure a sufficient number of correspondences.
Second, these poorly localized sampled points will result in inaccurate pose estimation. 
Third, non-salient points with indiscriminative descriptors degrade the inlier ratio.
Similarly, existing keypoint detectors trained independently cannot fully adapt to descriptors.

To this end, we design a tightly coupled joint keypoint detector and descriptor, \textit{i.e.}, TCKDD.
Inspired by KPConv\cite{thomas2019kpconv}, we propose a fully convolutional neural network for keypoint detection and description, which utilizes a KPConv-based encoder-decoder backbone for 3D feature embedding. In addition, it densely predicts both a descriptor and the saliency uncertainty via two independent point-wise MLP heads for each point.
The descriptors are learned through a quadruplet hardest contrastive loss in a self-supervised manner.
To fully exert the potential of descriptors, we quantitatively define a matchability index fully based on descriptors and design a novel probabilistic detection loss based on maximum likelihood estimation.
With this loss, the detection head can estimate the point-wise matchability robustly for keypoint selection and make the network concentrate on the learning of descriptors in salient regions.

\begin{figure}[t]
\centering
\includegraphics[width=\linewidth]{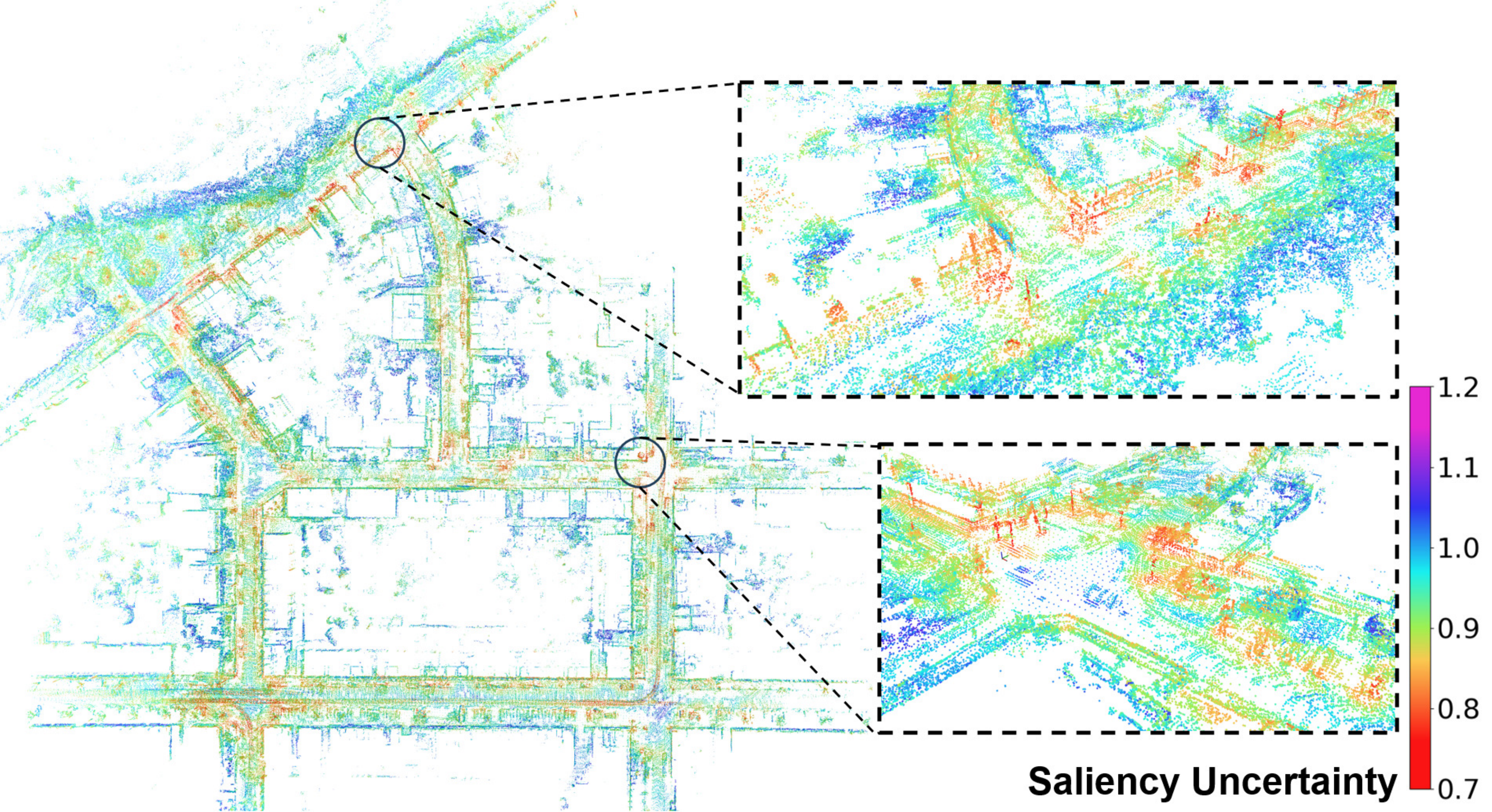}\vspace{-2mm}
\caption{The feature maps colored in saliency uncertainty built by our KDD-LOAM on KITTI sequence 07. Sharp corners and edges, distinguishable buildings, pillars, and vehicles are detected as salient regions (red), while flat surfaces, chaotic vegetation, and unstably scanned regions far from the sensor are detected as non-salient regions (blue). It is noteworthy that planar surfaces (most from roads) have been fitted as sparse surfels. }
\label{fig:map}
\vspace{-5mm}
\end{figure}

As a deep front-end, TCKDD can densely predict global and local context-aware descriptors and detect keypoints with higher matchability.
A series of experiments show that TCKDD achieves \textit{state-of-the-art} performance on both the indoor RGB-D camera dataset 3DMatch~\cite{zeng20173dmatch} and the outdoor LiDAR dataset KITTI~\cite{geiger2012kitti} for point cloud registration.
With this powerful front-end, we utilize RANSAC as the middle-end to establish sparse correspondences based on the descriptors and minimize the point-to-point metric at the back-end.
This pipeline can be operated in real-time for the registration between consecutive scans.
Consequently, we can design a keypoint detector and descriptors assisted LiDAR odometry and mapping framework, \textit{i.e.}, KDD-LOAM, with this keypoint descriptor matching-based registration pipeline as the LiDAR odometry module.
We propose to construct a voxel hash map consisting of only salient regions and fit the planar patches into sparse surfels, enabling a memory-efficient representation and fast nearest neighbor search.
Then the keypoints from the current scan will be aligned to the map based on both point-to-point and point-to-plane metrics for more accurate localization.
Experiments on the KITTI dataset demonstrate that the TCKDD-based odometry outperforms LOAM~\cite{zhang2014loam} which leverages the planar points and edge points for odometry by a large margin.
Without loop closure detection and pose graph optimization, KDD-LOAM achieves competitive performance in odometry while maintaining real-time performance.
Fig.~\ref{fig:map} is a demonstration of our keypoint detection and mapping results.
The main contributions of this work are summarized as follows:
\begin{itemize}
    \item We design a probabilistic detection loss to tightly couple the learning of 3D keypoint detection and description.
    \item We propose a keypoint detector and descriptors assisted real-time LiDAR odometry and mapping framework, which achieves more accurate and robust performance in odometry in comparison to LOAM.
    \item A series of experiments on both indoor and outdoor datasets show that our TCKDD achieves \textit{state-of-the-art} performance in 3D point cloud registration.
\end{itemize}

\section{Related Work}
\label{sec:related}
Feature matching is a prominent approach in point cloud registration, efficiently establishing reliable sparse correspondences based on descriptors. Early methods use local hand-crafted descriptors based on either histograms~\cite{rusu2008pfh, rusu2009fpfh} or signatures~\cite{salti2014shot}.
Recent focus has shifted to learning-based 3D descriptors. For instance, 3DMatch~\cite{zeng20173dmatch} and PerfectMatch~\cite{gojcic2019perfect} employ 3D CNNs to learn local volumetric descriptors, converting patches into truncated distance function (TDF) or smoothed density value (SDV) representations.
PPFNet~\cite{deng2018ppfnet} uses PointNet~\cite{qi2017pointnet} for global context-aware patch descriptors, while FCGF~\cite{choy2019fcgf} designs a sparse 3D convolutional encoder-decoder network.
SpinNet~\cite{ao2021spinnet} proposes a spatial point Transformer, converting point clouds as cylindrical volumes for transformation-invariant features.
Recent methods like~\cite{huang2021predator} learn point cloud interactions to enhance the inlier ratio, including coarse-to-fine registration approaches~\cite{yu2021cofinet, qin2023geotransformer} that achieve end-to-end correspondence learning.

Unlike the exploration of learning-based 3D descriptors, most 3D keypoint detectors are hand-crafted and target points with unique curvatures~\cite{chen2007det} or significant geometric variations in the principal direction~\cite{zhong2009iss} as keypoints.
However, they struggle with real-world scans that are noisy, sparse, and non-uniform.
Hence, researchers explore learning-based detectors for more reliable results.
USIP ~\cite{li2019usip} trains a feature proposal network via a probabilistic chamfer loss to predict 3D keypoints with high repeatability.
To adapt the keypoint detector to the descriptors, some researchers propose to jointly learn the 3D keypoint detector and descriptors.
3DFeat-Net~\cite{yew20183dfeat} designs a weakly supervised patch-wise network minimizing a saliency-weighted feature alignment triplet loss.
However, it does not explicitly prioritize keypoint detection performance.
D3Feat~\cite{bai2020d3feat} densely predicts descriptors with a fully convolutional encoder-decoder and obtains saliency scores from descriptors using a self-supervised detection loss.

LiDAR odometry estimation involves real-time point cloud registration typically based on ICP or NDT.
Nearly all modern SLAM systems are designed on top of odometry.
Zhang \textit{et al.}~\cite{zhang2014loam} propose LOAM that extracts and aligns planar and edge points to a sparse voxel grid-based feature map. 
LeGO-LOAM~\cite{shan2018lego} segments the point cloud to remove unstable parts and adds ground constraints to improve accuracy. 
Additionally, F-LOAM~\cite{wang2021floam} employs a faster non-iterative distortion compensation method to reduce the computational cost.
However, these methods rely on hand-crafted feature extraction, which is only suitable for small pose derivations and requires multiple iterations for reliable correspondences.

\section{Methodology}
In this work, we design a \textbf{t}ightly \textbf{c}oupled \textbf{k}eypoint \textbf{d}etector and \textbf{d}escriptor for 3D point cloud registration, \textit{i.e.}, TCKDD.
The principles of \textit{tight coupling} are three-fold: 1) the detector and the descriptor share a common feature extractor; 
2) the detector fully matches the matchability of the descriptors;
3) the detector and the descriptor can enhance each other via a probabilistic detection loss.
We further integrate TCKDD into a real-time LiDAR odometry and mapping system.

\subsection{Network Architecture}
We treat the joint learning of 3D keypoint detection and description as a multi-task learning paradigm and follow its fundamental neural network style, \textit{i.e.}, plugging several separated task-specific prediction heads into a shared feature extraction backbone.
Inspired by KPConv~\cite{thomas2019kpconv}, we propose a fully convolutional network for 3D keypoint detection and description.
KPConv directly operates on irregular point sets by interpolating point features to uniformly distributed kernel points for regular convolution, \textit{i.e.}, linear mapping, and summation of kernel responses.

\begin{figure}[t]
\centering
\includegraphics[width=\linewidth]{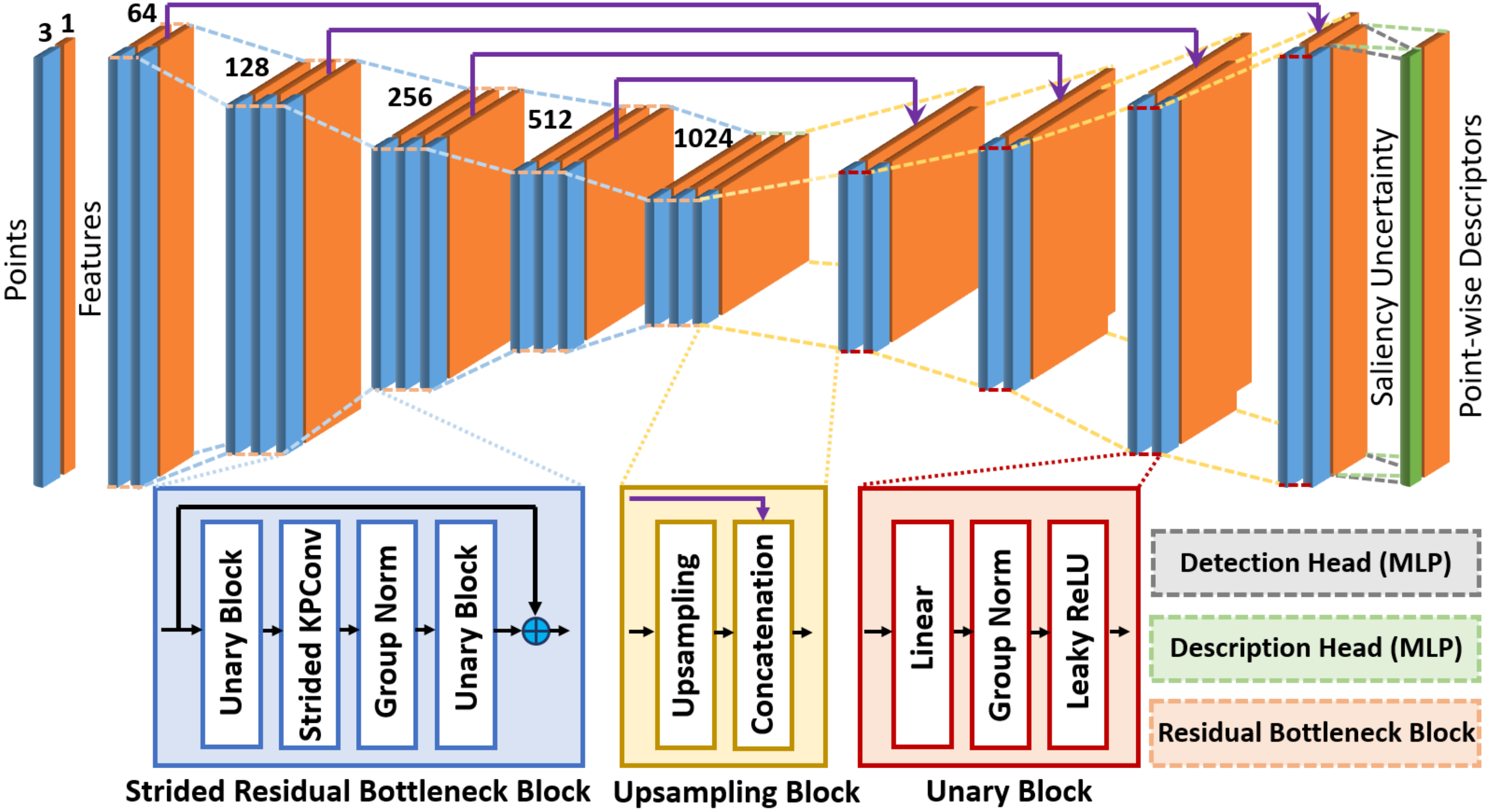}
\caption{The network architecture of TCKDD for jointly learning of 3D keypoint detection and description.}
\label{fig:network}
\vspace{-5mm}
\end{figure}

Utilizing the normalized KPConv, TCKDD can construct a fully convolutional network that directly consumes point sets, as depicted in Fig.~\ref{fig:network}.
The backbone can extract multi-scale 3D features at different encoder layers consisting of a stack of residual bottleneck blocks.
The locality of KPConv enables strided convolution for downsampling.
The decoder can recover the resolution via nearest upsampling and aggregate the multi-scale 3D features via skip connections and $1\times1$ convolution (unary blocks).
Different from the original KPFCNN~\cite{thomas2019kpconv}, we replace the batch normalization with group normalization~\cite{wu2018group}, which is robust \textit{w.r.t.} batch size and group-wise features.
Finally, TCKDD densely predicts both point-wise descriptors and saliency uncertainty based on the shared 3D features from the backbone via two independent all-MLP heads, \textit{i.e.}, $1\times1$ convolutional blocks.

\subsection{Quadruplet Contrastive Loss for Descriptor Learning}

Metric learning is widely used to train descriptors. Essentially it maps low-level geometric representations to a high-dimensional feature space, where descriptors of correctly associated point pairs are close, while those of other pairs differ by at least a margin.
The correspondence set $\mathbf{C}$ of two partially overlapped point clouds $P, Q$ is a set of the mutually nearest points $(p_i, q_j)$ satisfying $\Vert p_i - q_j \Vert_2 \leq R_p$.
The negative point set $\mathbf{N}_i$ of a point $p_i\in P$ is a set of points $q_k\in Q$ satisfying $\Vert p_i - q_k \Vert_2 \geq R_n (R_n\geq R_p)$.
Denote $d_i, d_j$ as the descriptors of $p_i\in P, q_j\in Q$, then the self-supervised descriptor loss can be designed as a hardest quadruplet contrastive loss according to metric learning strategies:
\begin{equation}
\begin{aligned}
\mathcal{L}&_{desc} = \frac{1}{\vert \mathbf{C} \vert}\sum_{(i,j)\in \mathbf{C}} \bigg\{ \lambda_p\left[D(d_i,d_j) - m_p\right]_{+}\\
&+ \left[m_n - \min_{k\in \mathbf{N}_i} D(d_i,d_k)\right]_{+}
+ \left[m_n - \min_{k\in \mathbf{N}_j} D(d_k,d_j)\right]_{+} \bigg\},
\label{eq:contrastive}
\end{aligned}
\end{equation}
where $D(\cdot,\cdot)$ is a distance metric and $m_p, m_n$ are positive and negative margins, respectively.
This loss can maximize the similarity between descriptors of true correspondences and minimize the maximum similarity otherwise.

\subsection{Self-supervised Probabilistic Detection Loss}
The principle of joint learning of 3D keypoint detection and description is to fully adapt the detector and the descriptors to each other.
Therefore, we treat the detection head of TCKDD as a saliency scoring network based on the matchability of descriptors.
According to (\ref{eq:contrastive}),
we can directly design a metric named \textit{matchability index} to characterize the matchability of a given descriptor $d_i$ quantitatively:
\begin{equation}
m_i = \left[D(d_i,d_j) - m_p\right]_{+} + \left[m_n - \min_{k\in \mathbf{N}_i} D(d_i,d_k)\right]_{+},
\label{eq:saliency}
\end{equation}
where $d_j$ is the descriptor of the correctly associated point of $p_i$ in point cloud $Q$.
This matchability index is actually the hardest triplet loss of a single descriptor in metric learning, which describes the distinctness of a descriptor in the feature space.
Particularly, we use the hardest negative pairs for negative mining so that the matchability index directly models the decision margin of descriptor matching and optimizes the decision boundary during metric learning.
A lower matchability index $m_i$ indicates greater matchability of the descriptor $d_i$, \textit{i.e.}, the 3D point $p_i$ is more salient.
Therefore, all we need is to learn a matchability index estimator as a keypoint detector.
We propose to learn a probabilistic model rather than a regressive model for matchability index estimation 
since the descriptor is predicted from a specific point cloud sampled from the surfaces of a dense 3D scene in a specific perspective, and so as the descriptors of the associated points.
Ideally, metric learning would construct a fully discriminative feature space where the matchability index of each descriptor is zero.
Hence, we choose an exponential distribution to model the matchability index $m_i$ with a parameter $\sigma_i$, or namely \textit{saliency uncertainty}:
\begin{equation}
p(m_i|\sigma_i) = \frac{1}{\sigma_i} \exp{\left( -\frac{m_i}{\sigma_i} \right)}.
\label{eq:probabilistic}
\end{equation}

It is notable that the detection head of our  TCKDD directly predicts point-wise saliency uncertainty as outputs.
Hence, we can design a probabilistic detection loss based on maximum likelihood estimation (MLE) to robustly fit this exponential distribution model:
\begin{equation}
\begin{aligned}
\mathcal{L}_{det} &= -\frac{1}{\vert \mathbf{C} \vert}\sum_{(i,j)\in \mathbf{C}} \left( \ln p(m_i|\sigma_i) + \ln p(m_j|\sigma_j) \right) \\
&= \frac{1}{\vert \mathbf{C} \vert}\sum_{(i,j)\in \mathbf{C}} \left( \ln \sigma_i + \frac{m_i}{\sigma_i} + \ln \sigma_j + \frac{m_j}{\sigma_j} \right).
\end{aligned}
\label{eq:det_loss}
\end{equation}
Theoretically, the first derivative of the log-likelihood indicates that the global optimality conditions for detector learning are $\sigma_i=m_i$, making the probabilistic detection loss effective for training the detection head as a robust matchability estimator.
Remarkably, this loss is also a weighted form of the hardest contrastive loss, with the descriptor losses of keypoints having higher weights than those of non-salient points.
When the detector meets global optimality, this detection loss turns out to be a logarithmic contrastive loss, prioritizing the descriptors of keypoints.
This approach enhances the matchability of keypoints, mitigating the negative effects of mining geometric features in non-salient areas like planar surfaces and disorganized regions.
Hence, this probabilistic detection loss can not only train a keypoint detector as a robust matchability estimator that fully accommodates the jointly learned descriptors but also promote the learning of keypoint descriptors via weighted metric learning.

\subsection{Keypoint Detector and Descriptors Assisted LOAM}
\begin{figure}[t]
\centering
\includegraphics[width=\linewidth]{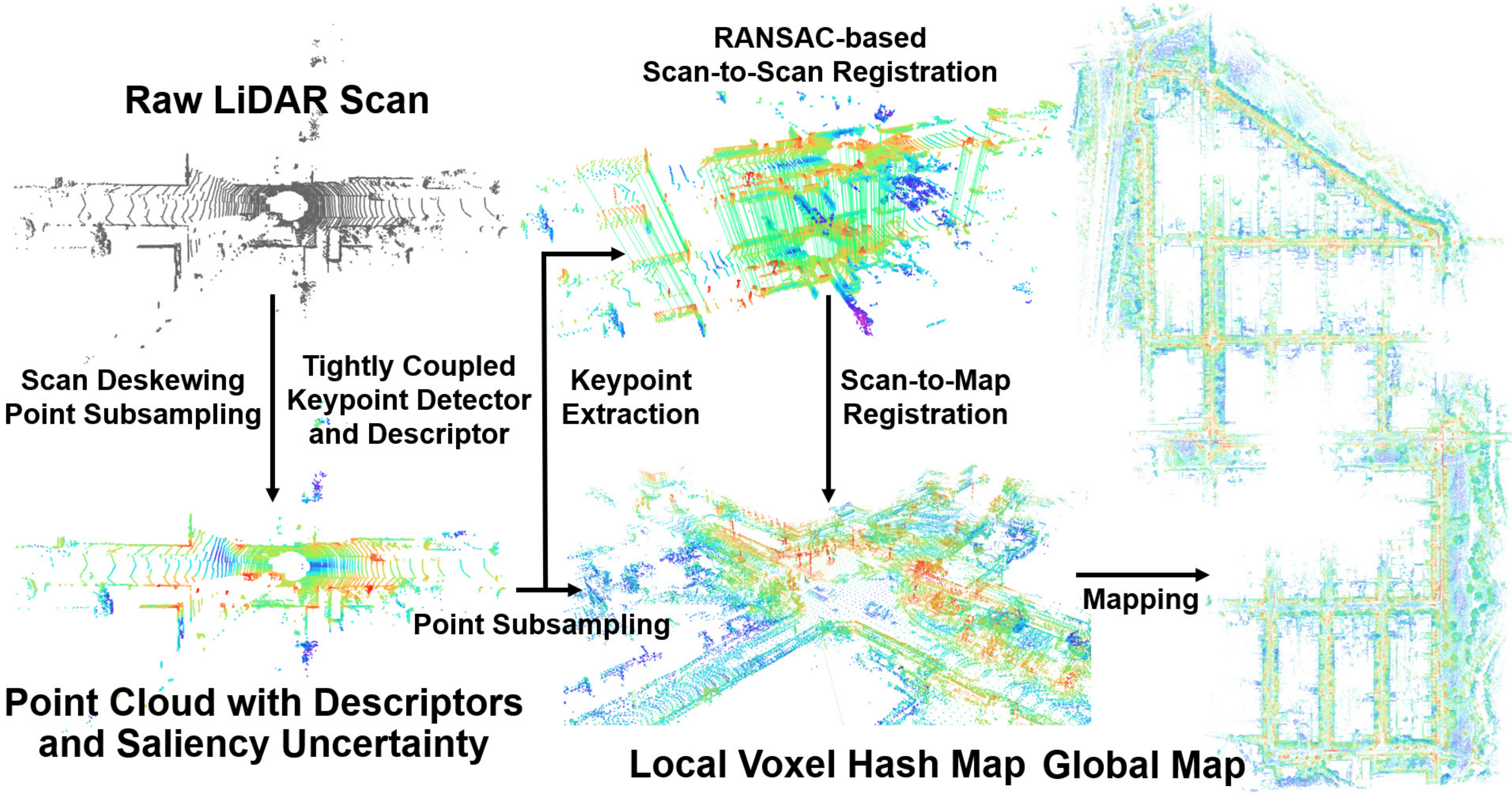}
\vspace{-5mm}
\caption{The system overview of KDD-LOAM.
We leverage the constant velocity model for scan deskewing and predict the point-wise descriptors and saliency uncertainty through TCKDD.
Based on a reliable relative pose guess from RANSAC-based scan-to-scan registration, KDD-LOAM achieves accurate odometry by aligning the deskewed and subsampled scan with a high-resolution yet memory-efficient local map.}
\label{fig:overview}
\vspace{-5mm}
\end{figure}
As illustrated in Fig.~\ref{fig:overview}, we integrate TCKDD into a LiDAR odometry and mapping system, \textit{i.e.}, KDD-LOAM, consisting of the following key components. 

\textbf{Scan deskewing and scan-to-scan registration.}
Similar to~\cite{vizzo2023kiss}, we first utilize the most generally applicable constant velocity model for scan deskewing, which requires no extra sensors involving time synchronization.
As a powerful front-end of point cloud registration, TCKDD can densely predict global and local context-aware descriptors and detect keypoints by sorting the saliency uncertainty.
We leverage RANSAC to establish sparse correspondences based on the descriptors and minimize the sum of point-to-point distances between inlier correspondences based on SVD.
This pipeline achieves real-time scan-to-scan registration to provide a solid initial guess for incremental ego-motion estimation.

\textbf{Keypoint subsampling and mapping.}
With a reliable relative pose guess from scan-to-scan registration, we refine the ego-pose estimate by aligning the deskewed and subsampled scan with the accumulated local map.
Unlike methods such as \cite{zhang2014loam} that create sparse feature maps of edge points or surface points with noisy directions or normals, we propose a voxel grid map capturing detailed geometry akin to surface reconstruction.
We keep it simple, approximating complex 3D surfaces in any topology with ample scan points.
Utilizing a voxel hash map with voxel size $v\times v\times v$ that stores up to $N_{max}$ points per voxel, we achieve efficient insertion, indexing, deletion, and nearest neighbor search compared to 3D arrays~\cite{zhang2014loam} or KD-trees~\cite{wang2021floam}.
Instead of high-resolution panoptic mapping~\cite{vizzo2023kiss}, KDD-LOAM accumulates a voxel hash map for only salient regions with low saliency uncertainty.
For better surface representation, we fit voxels with $N_{max}$ points to planes via least square regression.
Voxels meeting error criteria of regression are replaced with surfels represented by the point closest to their center, their normal vector, and radius $v$.
Our approach, in contrast to~\cite{vizzo2023kiss}, adapts voxel grids for adaptive 3D salient region reconstruction, combining dense points and sparse surfels for a memory-efficient representation applicable to global mapping.

Inspired by CT-ICP~\cite{dellenbach2022ct}, we adopt a two-stage voxel grid-based subsampling for sequential map update and ego-pose estimation.
In the first stage, we use voxel size $\alpha v$ ($\alpha\in(0,1]$) to downsample by reserving an original scan point per voxel to prevent discretization errors.
After scan-to-map registration, these subsampled points are transformed using the global pose estimate and added to the voxel hash map.
In the second stage, we propose a saliency-aware voxel grid subsampling using voxel size $\beta v$ ($\beta\in[1,2]$) to select keypoints for faster scan-to-map registration.
Non-salient points are discarded, while salient regions accommodate more points per voxel for accurate point cloud registration.

\textbf{Robust scan-to-map registration.}
We use scan-to-map registration for more accurate odometry as it proves more reliable and robust than scan-to-scan registration~\cite{zhang2014loam, behley2018suma}.
Our scan-to-map registration builds on the classic ICP algorithm, which typically establishes correspondences between two point clouds via nearest neighbor search.
With a reliable scan-to-scan relative pose guess grounded in geometrically consistent correspondences, it is more likely to avoid sub-optimal convergence and reduce ICP iterations.
However, ICP requires a hand-crafted maximum distance threshold for outlier rejection, which depends on the expected initial error.

To this end, we treat scan-to-map registration as compensation for scan-to-scan pose estimation, determining the maximum distance threshold by assessing deviations from the scan-to-scan pose estimate over time.
Denote this deviation as $\Delta T\in SE(3)$, the upper bound of the point deviation is
\begin{equation}
\delta(\Delta T)=\Vert \Delta t \Vert_2 + 2r\sin\left(\frac{1}{2} \arccos{\frac{tr(\Delta R)-1}{2}}\right),
\end{equation}
where $r$ is the maximum range of LiDAR scans, $\Delta R\in SO(3)$ and $\Delta t\in \mathbb{R}^3$ refer to the rotation and translation components of $\Delta T$, respectively.
Inspired by KISS-ICP~\cite{vizzo2023kiss}, we adopt a Gaussian distribution over $\delta(\Delta T)$ and compute its standard deviation $\sigma_t$ to robustly set the maximum distance threshold of ICP as the three-sigma bound $\tau_t = 3\sigma_t$.

Given a point from the current scan during data association, we first search the nearest point and the nearest surfel separately in the voxel hash map based on point-to-point distances.
Next, we evaluate the point-to-point distance from the nearest point and the point-to-plane distance from the nearest surfel to determine its correspondence.
Finally, the maximum distance threshold $\tau_t$ determines whether to accept this correspondence as an inlier.
This process establishes a set of point-to-point and point-to-plane correspondences for each ICP iteration.
A robust ego-pose (\textit{i.e.}, rotation $R\in SO(3)$ and translation $t\in\mathbb{R}^3$) is estimated by minimizing the sum of point-to-point residuals and point-to-plane residuals:
\begin{equation}
\min_{R,t}\sum_{p,q\in \mathcal{C}}\rho(e)=\frac{e(Rp+t,q)^2/2}{\sigma_t/3+e(Rp+t,q)^2},
\end{equation}
where $\rho$ is the Geman-McClure kernel with a strong outlier rejection property.
The optimal pose can be estimated via the Gauss-Newton method. The Jacobian can be derived by applying the left perturbation model with $\delta\xi=[\delta\rho^T\; \delta\phi^T]^T\in\mathbb{R}^6,\; \delta\xi^{\wedge}\in\mathfrak{se}(3)$.
For a point-to-point residual $e=\Vert Rp+t-q\Vert_2^2$, the Jacobian of $\mathbf{e}=Rp+t-q$ \textit{w.r.t.} $\delta\xi$ is
\begin{equation}
\label{eq:point-to-point}
\begin{aligned}
J_p &=\frac{\partial (Rp+t-q)}{\partial\delta\xi} = \lim_{\delta\xi\to 0}\frac{exp(\delta\phi^{\wedge})Rp+\delta\rho-Rp}{\delta\xi}\\
&=\begin{bmatrix} I_{3\times3} & -(Rp+t)^{\wedge}\end{bmatrix}_{3\times6}.
\end{aligned}
\end{equation}
For a point-to-plane residual $e=|n^T(Rp+t-q)|^2$, the Jacobian of $\mathbf{e}=nn^T(Rp+t-q)$ \textit{w.r.t.} $\delta\xi$ is derived as
\begin{equation}
J_s =\frac{\partial nn^T(Rp+t-q)}{\partial\delta\xi} = \begin{bmatrix} nn^T & -nn^T(Rp+t)^{\wedge}\end{bmatrix}.
\label{eq:point-to-plane}
\end{equation}
According to the chain rule of derivative, making the derivative of the objective function \textit{w.r.t.} the disturbance $\delta\xi$ equal to 0 will lead to the following equation:
\begin{equation}
\label{eq:gauss-newton}
\sum_i \frac{1}{(\sigma_t/3+e_i^2)^2} J_i^TJ_i\delta\xi = -\sum_i \frac{1}{(\sigma_t/3+e_i^2)^2}J^T_i\mathbf{e}_i.
\end{equation}
The registration is performed by repeating data association and solving (\ref{eq:gauss-newton}) until convergence.

\section{Experiments}
\label{sec:experiments}

In this section, we will evaluate our TCKDD regarding point cloud registration on both indoor (3DMatch~\cite{zeng20173dmatch}) and outdoor scenes (KITTI~\cite{geiger2012kitti}).
In addition, KDD-LOAM will be evaluated on the KITTI odometry benchmark against existing LiDAR-based odometry and SLAM systems.
Our source code will be released once the paper is accepted.

\subsection{Indoor Scenes: 3DMatch Benchmark}
3DMatch is a widely used 3D reconstruction benchmark including 62 indoor scenes collected by RGB-D cameras.
We use the training data preprocessed by~\cite{huang2021predator} and evaluate our TCKDD against both hand-crafted and learning-based descriptors on the official test set including scan pairs with $>30\%$ overlap using two metrics: feature matching recall (FMR)~\cite{deng2018ppfnet} and registration recall (RR)~\cite{zeng20173dmatch}.

\begin{table}
\scriptsize
\caption{Evaluation results on indoor datasets 3DMatch.}
\vspace{-1mm}
\label{table:3dmatch}
\centering
\setlength\tabcolsep{3pt}
\begin{tabular}{l|ccccc|ccccc}
\toprule
Sampled Points & 5000 & 2500 & 1000 & 500 & 250 & 5000 & 2500 & 1000 & 500 & 250\\
\cmidrule{1-11}
 & \multicolumn{5}{c|}{Feature Matching Recall (\%)} & \multicolumn{5}{c}{Registration Recall (\%)} \\
\cmidrule{1-11}
PerfectMatch~\cite{gojcic2019perfect} & 95.0 & 94.3 & 92.9 & 90.1 & 82.9
& 78.4 & 76.2 & 71.4 & 67.6 & 50.8\\
FCGF~\cite{choy2019fcgf} & 97.4 & 97.3 & 97.0 & 96.7 & 96.6
& 85.1 & 84.7 & 83.3 & 81.6 & 71.4\\
D3Feat~\cite{bai2020d3feat} & 95.6 & 95.4 & 94.5 & 94.1 & 93.1
& 81.6 & 84.5 & 83.4 & 82.4 & 77.9\\
SpinNet~\cite{ao2021spinnet} & 97.6 & 97.2 & 96.8 & 95.5 & 94.3
& 88.6 & 86.6 & 85.5 & 83.5 & 70.2\\
Predator~\cite{huang2021predator} & 96.6 & 96.6 & 96.5 & 96.3 & 96.5
& 89.0 & 89.9 & 90.6 & 88.5 & 86.6\\
YOHO~\cite{wang2022yoho} & \underline{98.2} & 97.6 & 97.5 & 97.7 & 96.0
& 90.8 & 90.3 & 89.1 & 88.6 & 84.5\\
CoFiNet~\cite{yu2021cofinet} & 98.1 & \textbf{98.3} & \textbf{98.1} & \textbf{98.2} & \textbf{98.3}
& 89.3 & 88.9 & 88.4 & 87.4 & 87.0\\
GeoTrans.~\cite{qin2023geotransformer} & 97.9 & 97.9 & 97.9 & \underline{97.9} & \underline{97.6}
& 92.0 & 91.8 & 91.8 & 91.4 & \textbf{91.2}\\
\cmidrule{1-11}
ours (rand) & \textbf{98.3} & \underline{98.0} & 97.9 & 97.7 & 96.8
& 91.9 & 91.6 & 91.3 & 88.4 & 81.5\\
ours (prob) & 98.1 & \underline{98.0} & 97.9 & 97.7 & 97.1 
& \textbf{93.0} & \underline{92.4} & \underline{92.0} & \underline{91.7} & 87.7\\
ours (NMS)  & 98.1 & 97.8 & \underline{98.0} & 97.8 & \underline{97.6}
& 92.1 & 91.5 & 91.3 & 91.3 & 89.7\\
ours (NMS-prob)& 98.1 & \underline{98.0} & 97.8 & 97.8 & 97.3
& \underline{92.4} & \textbf{93.1} & \textbf{92.7} & \textbf{92.4} & \underline{89.8}\\
\bottomrule
\end{tabular}
\vspace{-3mm}
\end{table}

TCKDD is evaluated with different numbers of keypoints in Table~\ref{table:3dmatch}, compared with \textit{state-of-the-art} learning-based descriptors PerfectMatch~\cite{gojcic2019perfect}, FCGF~\cite{choy2019fcgf}, D3Feat~\cite{bai2020d3feat}, SpinNet~\cite{ao2021spinnet}, Predator~\cite{huang2021predator}, YOHO~\cite{wang2022yoho} and coarse-to-fine registration methods CoFiNet~\cite{yu2021cofinet}, GeoTransformer~\cite{qin2023geotransformer}.
For TCKDD, we compare random sampling (rand) with three saliency-based keypoint selection strategies: probabilistic sampling (prob), non-maximum suppression (NMS), and probabilistic NMS (NMS-prob).
In terms of FMR, TCKDD without keypoints consistently outperforms all 3D descriptors and performs on par with GeoTransformer.
When sampled points are fewer than 1000, TCKDD with keypoints achieves higher FMR, showing more stable performance against existing descriptors.
In terms of RR, TCKDD with probabilistic keypoints outperforms all the descriptors and CoFiNet consistently by 2$\sim$39\%.
With over 250 sampled points, our probabilistic keypoints even surpass GeoTransformer notably.
Furthermore, the effectiveness and robustness of our keypoints are validated through consistent RR improvement, especially with fewer than 1000 sampled points.

\begin{figure}[t]
\centering
\includegraphics[width=\linewidth]{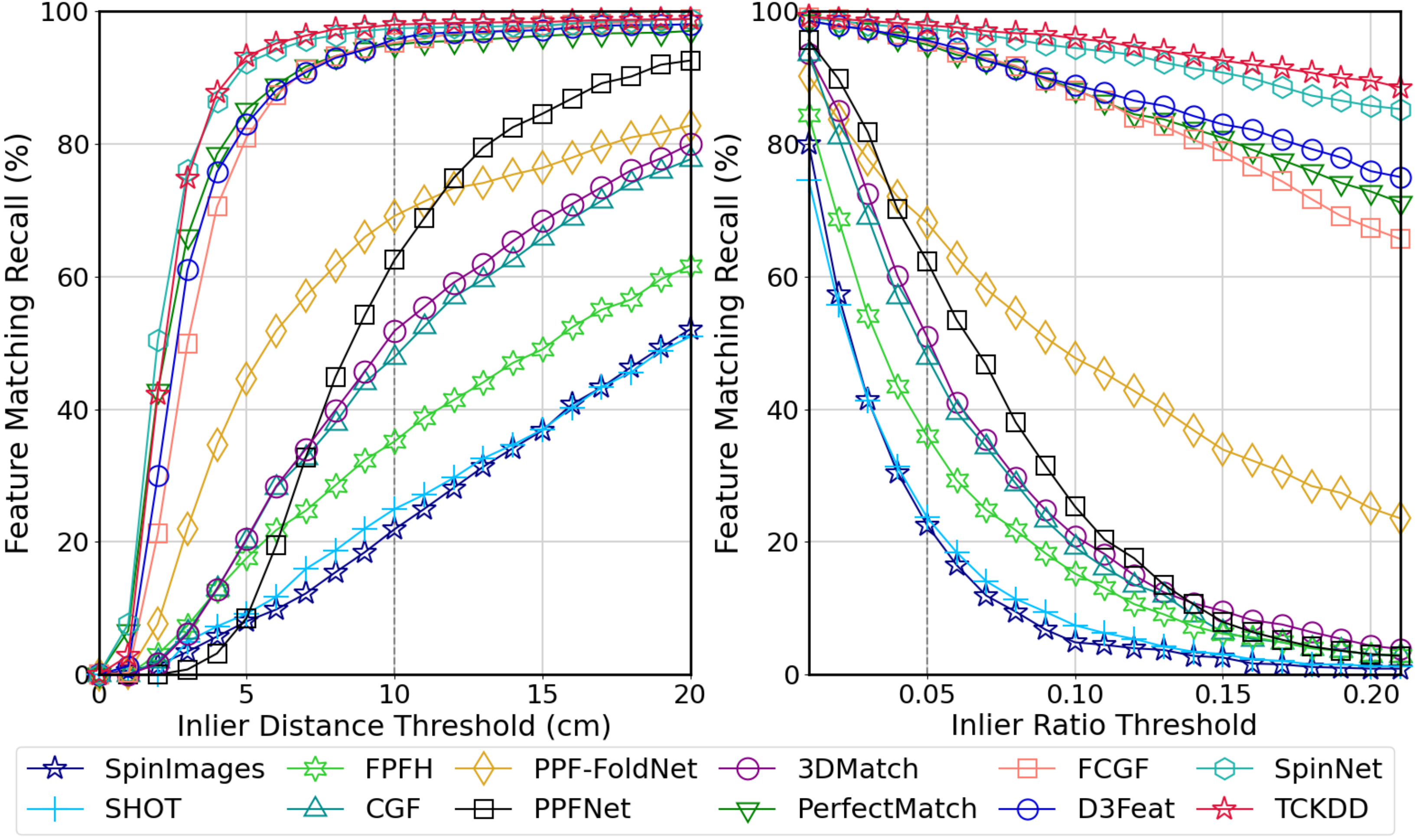}
\vspace{-5mm}
\caption{Feature matching recalls on the 3DMatch dataset in relation to inlier distance threshold $\tau_1$ (Left) and inlier ratio threshold $\tau_2$ (Right).}
\label{fig:fmr}
\end{figure}

We demonstrate the robustness of TCKDD (prob) by varying the inlier distance threshold $\tau_1$ and the inlier ratio threshold $\tau_2$ in FMR.
As shown in Fig.~\ref{fig:fmr}, we report the performance of hand-crafted descriptors SpinImages~\cite{johnson1999spin}, SHOT~\cite{salti2014shot}, FPFH~\cite{rusu2009fpfh} and ealier learning-based descriptors CGF~\cite{khoury2017cgf}, 3DMatch~\cite{zeng20173dmatch}, PPFNet~\cite{deng2018ppfnet}, PPF-FoldNet~\cite{deng2018ppffold}.
TCKDD consistently outperforms other methods with $\tau_1\geq5$cm and significantly surpasses them across all inlier ratio thresholds.
Under a stricter condition $\tau_2=0.2$, TCKDD maintains a high FMR of 89.3\%, while SpinNet, D3Feat, and FCGF drop to 85.7\%, 75.8\%, and 67.4\%, respectively, which highlights that TCKDD is more robust to maintain higher inlier ratio in challenging scenarios.

\subsection{Outdoor Scenes: KITTI Benchmark}
For the KITTI~\cite{geiger2012kitti} dataset, we use sequences 0 to 5 for training, 6 to 7 for validation, and 8 to 10 for testing.
We refine the GPS localization results via ICP~\cite{icp1992tpami} as ground truth.
Additionally, only point cloud pairs at least 10m away from each other are selected.
Following~\cite{huang2021predator}, we use three metrics for evaluation: \textit{relative translation error} (RTE), \textit{relative rotation error} (RRE) and \textit{registration recall} (RR).

\begin{table}
\footnotesize
\caption{Registration Performance on KITTI Odometry Dataset.}
\vspace{-1mm}
\label{tab:outdoor}
\centering
\begin{tabular}{p{93pt} | p{33pt}<{\centering} p{30pt}<{\centering} p{31pt}<{\centering}}
\toprule
Model & RTE (cm) & RRE (°) & RR (\%) \\
\cmidrule{1-4}
3DFeat-Net~\cite{yew20183dfeat} & 25.9 & 0.25 & 96.0\\
FCGF~\cite{choy2019fcgf} & 9.5 & 0.30 & 96.6\\
D3Feat~\cite{bai2020d3feat} & 7.2 & 0.30 & 99.8\\
SpinNet~\cite{ao2021spinnet} & 9.9 & 0.47 & 99.1\\
Predator~\cite{huang2021predator} & \textbf{6.8} & 0.27 & 99.8\\
CoFiNet~\cite{yu2021cofinet} & 8.2 & 0.41 & 99.8\\
GeoTransformer (RANSAC) & 7.4 & 0.27 & 99.8\\
GeoTransformer (LGR)~\cite{qin2023geotransformer} & \textbf{6.8} & \textbf{0.24} & 99.8\\
TCKDD (ours, prob) & \textbf{6.8} & 0.27 & \textbf{100.0} \\
\bottomrule
\end{tabular}
\vspace{-3mm}
\end{table}

In Table~\ref{tab:outdoor}, TCKDD is compared with the \textit{state-of-the-art} descriptors 3DFeat-Net~\cite{yew20183dfeat}, FCGF~\cite{choy2019fcgf}, D3Feat~\cite{bai2020d3feat}, SpinNet~\cite{ao2021spinnet}, Predator~\cite{huang2021predator} and coarse-to-fine registration methods CoFiNet~\cite{yu2021cofinet}, GeoTransformer~\cite{qin2023geotransformer}.
TCKDD achieves \textit{state-of-the-art} RTE and the highest registration recall of 100\%, which demonstrates the effectiveness and robustness.
Within TCKDD, we compare random sampling (rand) with two saliency-based keypoint selection strategies: sorting (sort) and probabilistic sampling (prob).
As shown in Table~\ref{tab:outabl}, TCKDD maintains 100\% of RR even with only 1000 randomly sampled points, indicating the robustness of its descriptors in establishing sparse yet reliable correspondences.
The sorted keypoints notably reduce RTE compared to random sampling, especially with fewer sampled points for registration, thus underscoring TCKDD's ability to detect keypoints with high matchability and repeatability.

\begin{table}
\footnotesize
\caption{Ablation Studies of Keypoint Detection on KITTI Dataset.}
\vspace{-1mm}
\label{tab:outabl}
\centering
\begin{tabular}{p{34pt}<{\centering} | p{55pt}<{\centering} p{55pt}<{\centering} p{55pt}<{\centering}}
\toprule
Points & 5000 & 2500 & 1000 \\
\cmidrule{1-4}
 & \multicolumn{3}{c}{RTE (cm) / RRE (°) / RR (\%)} \\
\cmidrule{1-4}
ours\;(rand) & 7.1\;/\;0.26\;/\;100.0 & 8.4\;/\;0.30\;/\;100.0 & 14.6\;/\;0.50\;/\;100.0 \\
ours\;(sort) & 6.9\;/\;0.30\;/\;100.0 & 7.6\;/\;0.39\;/\;100.0 & 
\;\;8.9\;/\;0.55\;/\;100.0 \\
ours\;(prob) & 6.8\;/\;0.27\;/\;100.0 & 7.6\;/\;0.33\;/\;100.0 & 10.3\;/\;0.46\;/\;100.0 \\
\bottomrule
\end{tabular}
\end{table}

\subsection{Evaluation of LiDAR Odometry and Mapping Systems}

In this subsection, we design experiments to demonstrate that
1) TCKDD effectively improves odometry accuracy against its baseline A-LOAM; 
2) KDD-LOAM significantly reduces cumulative error in scan-to-scan registration and outperforms classic LiDAR odometry or SLAM systems;
3) KDD-LOAM achieves more memory-efficient mapping while maintaining performance comparable to KISS-ICP.
All the algorithms are evaluated with the mean relative pose error (RPE) over trajectories of 100 to 800m (relative translation error in \% / relative rotational error in °/100m)~\cite{geiger2012kitti}.
Sequence 08 is excluded from evaluation due to significant errors in its ground-truth localization results.

\begin{table}
\footnotesize
\caption{Ablation studies of LiDAR odometry on KITTI benchmark.}
\vspace{-1mm}
\label{tab:loam}
\centering
\begin{tabular}{p{17pt}<{\centering} | p{40pt}<{\centering} p{40pt}<{\centering} | p{40pt}<{\centering} p{40pt}<{\centering}}
\toprule
\multirow{2}{*}{Seq} & \multicolumn{2}{c|}{A-LOAM} & \multicolumn{2}{c}{TCKDD + A-LOAM}\\
 & scan-to-scan & scan-to-map & scan-to-scan & scan-to-map\\
\cmidrule{1-5}
00 & 4.13 / 1.72 & 0.81 / 0.31 & \textbf{2.28} / \textbf{0.98} & \textbf{0.67} / \textbf{0.26} \\
01 & 3.46 / 0.95 & \textbf{2.01} / 0.52 & \textbf{2.89} / \textbf{0.80} & 2.19 / \textbf{0.48} \\
02 & 7.47 / 2.55 & 4.66 / 1.46 & \textbf{2.03} / \textbf{0.91} & \textbf{0.99} / \textbf{0.35} \\
03 & 4.35 / 2.08 & 0.92 / 0.47 & \textbf{2.03} / \textbf{1.40} & \textbf{0.90} / \textbf{0.44} \\
04 & 1.65 / 0.81 & 0.72 / 0.36 & \textbf{0.67} / \textbf{0.49} & \textbf{0.62} / \textbf{0.29} \\
05 & 4.06 / 1.66 & 0.51 / 0.24 & \textbf{1.96} / \textbf{1.00} & \textbf{0.45} / \textbf{0.22} \\
06 & \textbf{1.11} / \textbf{0.51} & \textbf{0.59} / \textbf{0.27} & 2.01 / 1.27 & \textbf{0.59} / \textbf{0.27} \\
07 & 2.84 / 1.80 & 0.44 / 0.24 & \textbf{1.80} / \textbf{1.36} & \textbf{0.43} / \textbf{0.22} \\
09 & 5.75 / 1.88 & 0.70 / 0.30 & \textbf{3.05} / \textbf{1.28} & \textbf{0.62} / \textbf{0.23} \\
10 & 3.60 / 1.76 & 0.98 / 0.38 & \textbf{3.26} / \textbf{1.38} & \textbf{0.84} / \textbf{0.34} \\
\bottomrule
\end{tabular}
\vspace{-3mm}
\end{table}

We first demonstrate the effectiveness of TCKDD by replacing the scan-to-scan registration step of A-LOAM with TCKDD-based RANSAC. 
As shown in Table~\ref{tab:loam}, TCKDD-based scan-to-scan registration significantly outperforms A-LOAM based on hand-crafted keypoints, providing a much more reliable and accurate pose guess for the subsequent scan-to-map registration.
Furthermore, we integrate TCKDD-based scan-to-scan registration with the subsequent mapping step of A-LOAM.
Apart from sequence 01 collected from a featureless environment, TCKDD-aided A-LOAM outperforms A-LOAM by a large margin, especially in sequences 00, 02, and 10 with 0.14\%, 3.67\% and 0.14\% improvements of average RTEs, respectively.
These results indicate that TCKDD-based odometry outperforms hand-crafted features and effectively complements existing mapping approaches.

\begin{table}
\scriptsize
\setlength\tabcolsep{3.8pt}
\caption{Relative pose errors of LiDAR odometry on KITTI dataset.}
\vspace{-1mm}
\label{tab:kddloam}
\centering
\begin{tabular}{c|cccccc}
\toprule
Seq & LOAM & F-LOAM & SuMa & SuMa++ & KISS-ICP & KDD-LOAM\\
\cmidrule{1-7}
00 & 0.78\;/\;- & 0.92\;/\;0.43 & 0.77\;/\;0.32 & 0.65\;/\;0.22 & \textbf{0.52}\;/\;0.19 & \textbf{0.52}\;/\;\textbf{0.18}\\
01 & 1.43\;/\;- & 2.80\;/\;0.60 & 11.15\;/\;0.76 & 1.63\;/\;0.47 & \textbf{0.65}\;/\;\textbf{0.14} & 0.76\;/\;\textbf{0.14}\\
02 & 0.92\;/\;- & 1.56\;/\;0.52 & 2.93\;/\;0.93 & 3.54\;/\;0.14 & 0.53\;/\;0.15 & \textbf{0.51}\;/\;\textbf{0.14} \\
03 & 0.86\;/\;- & 1.09\;/\;0.66 & 1.25\;/\;0.61 & 0.67\;/\;0.47 & \textbf{0.66}\;/\;\textbf{0.16} & 0.67\;/\;\textbf{0.16} \\
04 & 0.71\;/\;- & 1.43\;/\;0.52 & 0.86\;/\;0.27 & \textbf{0.34}\;/\;0.27 & 0.35\;/\;0.13 & 0.37\;/\;\textbf{0.07} \\
05 & 0.57\;/\;- & 0.79\;/\;0.36 & 0.56\;/\;0.32 & 0.40\;/\;0.19 & 0.32\;/\;0.14 & \textbf{0.26}\;/\;\textbf{0.12} \\
06 & 0.65\;/\;- & 0.72\;/\;0.39 & 0.64\;/\;0.51 & 0.47\;/\;0.27 & \textbf{0.26}\;/\;\textbf{0.08} & \textbf{0.26}\;/\;\textbf{0.08} \\
07 & 0.63\;/\;- & 0.54\;/\;0.39 & 0.47\;/\;0.37 & 0.39\;/\;0.28 & 0.32\;/\;0.16 & \textbf{0.31}\;/\;\textbf{0.15} \\
09 & 0.77\;/\;- & 1.28\;/\;0.55 & 0.79\;/\;0.41 & 0.58\;/\;0.20 & \textbf{0.48}\;/\;0.13 & 0.50\;/\;\textbf{0.12} \\
10 & 0.79\;/\;- & 1.77\;/\;0.58 & 0.99\;/\;0.44 & 0.67\;/\;0.30 & 0.60\;/\;0.20 & \textbf{0.53}\;/\;\textbf{0.17}\\
\cmidrule{1-7}
Avg & 0.81\;/\;- & 1.29\;/\;0.50 & 2.04\;/\;0.49 & 0.93\;/\;0.28 & \textbf{0.47}\;/\;0.15 & \textbf{0.47}\;/\;\textbf{0.13}\\
Avg$^\dag$ & 0.74\;/\;- & 1.12\;/\;0.49 & 1.03\;/\;0.46 & 0.86\;/\;0.26 & 0.45\;/\;0.15 & \textbf{0.44}\;/\;\textbf{0.13} \\
\bottomrule
\end{tabular}
\end{table}

Finally, we evaluate KDD-LOAM against \textit{state-of-the-art} LiDAR odometry methods on the KITTI dataset~\cite{geiger2012kitti}, including LOAM~\cite{zhang2014loam}, F-LOAM~\cite{wang2021floam}, SuMa~\cite{behley2018suma}, SuMa++~\cite{chen2019suma++}, and KISS-ICP~\cite{vizzo2023kiss}.
As shown in Table~\ref{tab:kddloam}, KDD-LOAM consistently outperforms classic odometry systems, LOAM, F-LOAM, and SLAM systems SuMa, SuMa++ across nearly all the scenes. This underscores the effectiveness and robustness of KDD-LOAM.
Compared with KISS-ICP using a similar voxel hash map and an ICP-based scan-to-map registration step with adaptive thresholds, KDD-LOAM achieves lower RREs in all sequences while performing on par with KISS-ICP in RTEs, showcasing its potential to achieve lower global localization drifts.
The average RPEs are reported as Avg, while Avg$^\dag$ stands for the results without the featureless sequence 01.
Except for sequence 01, KDD-LOAM even achieves a lower average RPE than KISS-ICP with a more memory-efficient map representation.
We compare the average memory usage for local maps of some long sequences in Table~\ref{tab:memory}, which indicates that KDD-LOAM consumes 16.6\% less memory for local mapping than KISS-ICP.

\begin{table}
\footnotesize
\setlength\tabcolsep{3.8pt}
\caption{Memory Usage for Local Mapping of Different Methods (KB).}
\vspace{-1mm}
\label{tab:memory}
\centering
\begin{tabular}{c|cccccc}
\toprule
Sequence & 00 & 02 & 05 & 09 & 10 & Avg\\
\cmidrule{1-7}
KISS-ICP & 4685.6 & 3936.6 & 5038.9 & 4785.9 & 3836.8 & 4456.8 \\
KDD-LOAM & 4065.5 & 3091.2 & 4434.4 & 3881.8 & 3101.7 & 3714.9 \\
\bottomrule
\end{tabular}
\vspace{-3mm}
\end{table}

\section{Conclusions}
This paper presents a tightly coupled 3D keypoint detector and descriptor for point cloud registration along with a keypoint detector and descriptor assisted LiDAR odometry and mapping system.
We exploit the multi-task learning paradigm with a carefully designed probabilistic detection loss to learn a fully convolutional 3D descriptor and a keypoint detector fully adapted to it.
Our odometry and mapping system achieves robust registration and memory-efficient mapping based on dense keypoints and sparse surfels.
The evaluation results indicate that our keypoint detector and descriptor are robust to different range-sensing technologies and achieve \textit{state-of-the-art} registration recall.
The experiments on the KITTI benchmark demonstrate that our real-time and memory-efficient KDD-LOAM performs on par with \textit{state-of-the-art} LiDAR odometry systems.
In future work, we intend to investigate keypoint descriptor-based loop closure detection and pose graph optimization to extend our KDD-LOAM to a full SLAM system.

{\small
\bibliographystyle{ieeetr}
\bibliography{kddloam}

\begin{thebibliography}{10}

\bibitem{behley2018suma}
J.~Behley and C.~Stachniss, ``Efficient surfel-based slam using 3d laser range data in urban environments.,'' in {\em Robotics: Science and Systems}, vol.~2018, p.~59, 2018.

\bibitem{zeng20173dmatch}
A.~Zeng, S.~Song, M.~Nie{\ss}ner, M.~Fisher, J.~Xiao, and T.~Funkhouser, ``3dmatch: Learning local geometric descriptors from rgb-d reconstructions,'' in {\em Proceedings of the IEEE Conference on Computer Vision and Pattern Recognition}, pp.~1802--1811, 2017.

\bibitem{icp1992tpami}
P.~Besl and N.~D. McKay, ``A method for registration of 3-d shapes,'' {\em IEEE Transactions on Pattern Analysis and Machine Intelligence}, vol.~14, no.~2, pp.~239--256, 1992.

\bibitem{chen2007det}
H.~Chen and B.~Bhanu, ``3d free-form object recognition in range images using local surface patches,'' {\em Pattern Recognition Letters}, vol.~28, no.~10, pp.~1252--1262, 2007.

\bibitem{zhong2009iss}
Y.~Zhong, ``Intrinsic shape signatures: A shape descriptor for 3d object recognition,'' in {\em IEEE International conference on Computer Vision Workshops, ICCV workshops}, pp.~689--696, IEEE, 2009.

\bibitem{rusu2008pfh}
R.~B. Rusu, N.~Blodow, Z.~C. Marton, and M.~Beetz, ``Aligning point cloud views using persistent feature histograms,'' in {\em 2008 IEEE/RSJ International Conference on Intelligent Robots and Systems}, pp.~3384--3391, IEEE, 2008.

\bibitem{rusu2009fpfh}
R.~B. Rusu, N.~Blodow, and M.~Beetz, ``Fast point feature histograms (fpfh) for 3d registration,'' in {\em 2009 IEEE International Conference on Robotics and Automation}, pp.~3212--3217, IEEE, 2009.

\bibitem{salti2014shot}
S.~Salti, F.~Tombari, and L.~Di~Stefano, ``Shot: Unique signatures of histograms for surface and texture description,'' {\em Computer Vision and Image Understanding}, vol.~125, pp.~251--264, 2014.

\bibitem{bai2020d3feat}
X.~Bai, Z.~Luo, L.~Zhou, H.~Fu, L.~Quan, and C.-L. Tai, ``D3feat: Joint learning of dense detection and description of 3d local features,'' in {\em Proceedings of the IEEE/CVF Conference on Computer Vision and Pattern Recognition}, pp.~6359--6367, 2020.

\bibitem{thomas2019kpconv}
H.~Thomas, C.~R. Qi, J.-E. Deschaud, B.~Marcotegui, F.~Goulette, and L.~J. Guibas, ``Kpconv: Flexible and deformable convolution for point clouds,'' in {\em Proceedings of the IEEE/CVF International Conference on Computer Vision}, pp.~6411--6420, 2019.

\bibitem{geiger2012kitti}
A.~Geiger, P.~Lenz, and R.~Urtasun, ``Are we ready for autonomous driving? the kitti vision benchmark suite,'' in {\em IEEE Conference on Computer Vision and Pattern Recognition}, pp.~3354--3361, IEEE, 2012.

\bibitem{zhang2014loam}
J.~Zhang and S.~Singh, ``Loam: Lidar odometry and mapping in real-time.,'' in {\em Robotics: Science and systems}, 2014.

\bibitem{gojcic2019perfect}
Z.~Gojcic, C.~Zhou, J.~D. Wegner, and A.~Wieser, ``The perfect match: 3d point cloud matching with smoothed densities,'' in {\em Proceedings of the IEEE/CVF Conference on Computer Vision and Pattern Recognition}, pp.~5545--5554, 2019.

\bibitem{deng2018ppfnet}
H.~Deng, T.~Birdal, and S.~Ilic, ``Ppfnet: Global context aware local features for robust 3d point matching,'' in {\em Proceedings of the IEEE Conference on Computer Vision and Pattern Recognition}, pp.~195--205, 2018.

\bibitem{qi2017pointnet}
C.~R. Qi, H.~Su, K.~Mo, and L.~J. Guibas, ``Pointnet: Deep learning on point sets for 3d classification and segmentation,'' in {\em Proceedings of the IEEE Conference on Computer Vision and Pattern Recognition}, pp.~652--660, 2017.

\bibitem{choy2019fcgf}
C.~Choy, J.~Park, and V.~Koltun, ``Fully convolutional geometric features,'' in {\em Proceedings of the IEEE/CVF International Conference on Computer Vision}, pp.~8958--8966, 2019.

\bibitem{ao2021spinnet}
S.~Ao, Q.~Hu, B.~Yang, A.~Markham, and Y.~Guo, ``Spinnet: Learning a general surface descriptor for 3d point cloud registration,'' in {\em Proceedings of the IEEE/CVF Conference on Computer Vision and Pattern Recognition}, pp.~11753--11762, 2021.

\bibitem{huang2021predator}
S.~Huang, Z.~Gojcic, M.~Usvyatsov, A.~Wieser, and K.~Schindler, ``Predator: Registration of 3d point clouds with low overlap,'' in {\em Proceedings of the IEEE/CVF Conference on Computer Vision and Pattern Recognition}, pp.~4267--4276, 2021.

\bibitem{yu2021cofinet}
H.~Yu, F.~Li, M.~Saleh, B.~Busam, and S.~Ilic, ``Cofinet: Reliable coarse-to-fine correspondences for robust point cloud registration,'' {\em Advances in Neural Information Processing Systems}, vol.~34, pp.~23872--23884, 2021.

\bibitem{qin2023geotransformer}
Z.~Qin, H.~Yu, C.~Wang, Y.~Guo, Y.~Peng, S.~Ilic, D.~Hu, and K.~Xu, ``Geotransformer: Fast and robust point cloud registration with geometric transformer,'' {\em IEEE Transactions on Pattern Analysis and Machine Intelligence}, 2023.

\bibitem{li2019usip}
J.~Li and G.~H. Lee, ``Usip: Unsupervised stable interest point detection from 3d point clouds,'' in {\em Proceedings of the IEEE/CVF International Conference on Computer Vision}, pp.~361--370, 2019.

\bibitem{yew20183dfeat}
Z.~J. Yew and G.~H. Lee, ``3dfeat-net: Weakly supervised local 3d features for point cloud registration,'' in {\em Proceedings of the European Conference on Computer Vision (ECCV)}, pp.~607--623, 2018.

\bibitem{shan2018lego}
T.~Shan and B.~Englot, ``Lego-loam: Lightweight and ground-optimized lidar odometry and mapping on variable terrain,'' in {\em 2018 IEEE/RSJ International Conference on Intelligent Robots and Systems (IROS)}, pp.~4758--4765, IEEE, 2018.

\bibitem{wang2021floam}
H.~Wang, C.~Wang, C.-L. Chen, and L.~Xie, ``F-loam: Fast lidar odometry and mapping,'' in {\em 2021 IEEE/RSJ International Conference on Intelligent Robots and Systems (IROS)}, pp.~4390--4396, IEEE, 2021.

\bibitem{wu2018group}
Y.~Wu and K.~He, ``Group normalization,'' in {\em Proceedings of the European Conference on Computer Vision (ECCV)}, pp.~3--19, 2018.

\bibitem{vizzo2023kiss}
I.~Vizzo, T.~Guadagnino, B.~Mersch, L.~Wiesmann, J.~Behley, and C.~Stachniss, ``Kiss-icp: In defense of point-to-point icp--simple, accurate, and robust registration if done the right way,'' {\em IEEE Robotics and Automation Letters}, vol.~8, no.~2, pp.~1029--1036, 2023.

\bibitem{dellenbach2022ct}
P.~Dellenbach, J.-E. Deschaud, B.~Jacquet, and F.~Goulette, ``Ct-icp: Real-time elastic lidar odometry with loop closure,'' in {\em IEEE International Conference on Robotics and Automation (ICRA)}, pp.~5580--5586, IEEE, 2022.

\bibitem{wang2022yoho}
H.~Wang, Y.~Liu, Z.~Dong, and W.~Wang, ``You only hypothesize once: Point cloud registration with rotation-equivariant descriptors,'' in {\em Proceedings of the 30th ACM International Conference on Multimedia}, pp.~1630--1641, 2022.

\bibitem{johnson1999spin}
A.~E. Johnson and M.~Hebert, ``Using spin images for efficient object recognition in cluttered 3d scenes,'' {\em IEEE Transactions on Pattern Analysis and Machine Intelligence}, vol.~21, no.~5, pp.~433--449, 1999.

\bibitem{khoury2017cgf}
M.~Khoury, Q.-Y. Zhou, and V.~Koltun, ``Learning compact geometric features,'' in {\em Proceedings of the IEEE International Conference on Computer Vision}, pp.~153--161, 2017.

\bibitem{deng2018ppffold}
H.~Deng, T.~Birdal, and S.~Ilic, ``Ppf-foldnet: Unsupervised learning of rotation invariant 3d local descriptors,'' in {\em Proceedings of the European Conference on Computer Vision (ECCV)}, pp.~602--618, 2018.

\bibitem{chen2019suma++}
X.~Chen, A.~Milioto, E.~Palazzolo, P.~Giguere, J.~Behley, and C.~Stachniss, ``Suma++: Efficient lidar-based semantic slam,'' in {\em 2019 IEEE/RSJ International Conference on Intelligent Robots and Systems (IROS)}, pp.~4530--4537, IEEE, 2019.

\end{thebibliography}
}

\end{document}